\documentclass{article}
\usepackage{iclr2025_conference,times}

\usepackage{amsmath,amsfonts,bm}

\def\eqref#1{equation~\ref{#1}}

\def\1{\bm{1}}

\DeclareMathAlphabet{\mathsfit}{\encodingdefault}{\sfdefault}{m}{sl}
\SetMathAlphabet{\mathsfit}{bold}{\encodingdefault}{\sfdefault}{bx}{n}

\usepackage[pagebackref,breaklinks,colorlinks,citecolor=blue]{hyperref}

\usepackage{url}

\usepackage{placeins}
\usepackage{wrapfig}
\usepackage{makecell}
\usepackage{booktabs}
\usepackage{arydshln}
\usepackage{pifont}
\usepackage{array}

\usepackage{tcolorbox}
\usepackage{makecell}
\usepackage{url}
\usepackage{multirow}
\usepackage{marvosym}

\usepackage{listings}

\usepackage{xcolor}

\definecolor{added}{RGB}{34,139,34}
\definecolor{deleted}{RGB}{220,20,60}
\definecolor{comment}{RGB}{0,102,204}

\newcommand{\added}[1]{\textcolor{black}{#1}}

\newcommand{\citeps}[1]{{\citep{#1}}}

\newcommand{\red}[1]{{\color{black} {\textbf{{#1}}}}}
\newcommand{\gre}[1]{{\color{black} {\textbf{{{#1}}}}}}

\newcommand{\blu}[1]{{ {{#1}}}}

\newcommand{\eg}{\textit{e.g.}}

\title{RGB-Event ISP: The Dataset and Benchmark}

\iclrfinalcopy
\author{Yunfan Lu, Yanlin Qian, Ziyang Rao, Junren Xiao, Liming Chen$^2$, Hui Xiong\thanks{corresponding author} \\AI Thrust, HKUST(GZ);~~AlpsenTek$^2$\\
{\small\texttt{ylu066@connect.hkust-gz.edu.cn, qianyanlin619812051@gmail.com}} \\
{\small\texttt{\{zrao538,jxiao767\}@connect.hkust-gz.edu.cn}} \\
{\small\texttt{liming.shen@alpsentek.com, xionghui@ust.hk}}
}

\begin{document}

\maketitle

\begin{abstract}
Event-guided imaging has received significant attention due to its potential to revolutionize instant imaging systems.
However, the prior methods primarily focus on enhancing RGB images in a post-processing manner,  neglecting the challenges of image signal processor (ISP) dealing with event sensor and the benefits events provide for reforming the ISP process.
To achieve this, we conduct the first research on event-guided ISP.
First, we present a new event-RAW paired dataset, collected with a novel but still confidential sensor that records \textbf{pixel-level aligned} events and RAW images. This dataset includes 3373 RAW images with $2248\times 3264$ resolution and their corresponding events, spanning 24 scenes with 3 exposure modes and 3 lenses.
Second, we propose a conventional ISP pipeline to generate good RGB frames as reference.
This conventional ISP pipleline performs basic ISP operations, \eg demosaicing, white balancing, denoising and color space transforming, with a ColorChecker as reference.
Third, we classify the existing learnable ISP methods into 3 classes, and select multiple methods to train and evaluate on our new dataset.
Lastly, since there is no prior work for reference, we propose a simple event-guided ISP method and test it on our dataset.
We further put forward key technical challenges and future directions in RGB-Event ISP.
In summary, to the best of our knowledge, this is the very first research focusing on event-guided ISP, and we hope it will inspire the community.
The code and dataset are available at: \url{https://github.com/yunfanLu/RGB-Event-ISP}.
\end{abstract}

\section{Introduction}
Since their invention in 1975, digital cameras have profoundly impacted various aspects of modern society~\citeps{delbracio2021mobile,kyung2016theory}.
Active pixel sensors (APS)~\citeps{liebe1998active} are used as the core of cameras to capture RGB color signals, recording images or videos.
This technology forms the foundation for widespread applications in smartphones~\citeps{delbracio2021mobile}, autopilot systems~\citeps{ingle2016tesla}, drones~\citeps{zhu2018vision}, virtual reality~\citeps{huang20176}, and more.
However, nowadays APS has reached a bottleneck \textit{wrt.} power consumption, frame rate, and dynamic range due to its global recording characteristics~\citeps{gallego2020event}.
Event vision sensors (EVS), with their inherent asynchronous recording property, achieve lower power consumption ($<10mW$), lower latency ($<1ms$), and higher dynamic range ($>120dB$) \citeps{gallego2020event}.
As a result, integrating EVS as a significant enhancement to APS imaging system has received considerable attention in recent years~\citeps{lu2023learning,tulyakov2021time,gallego2020event,tulyakov2022time}.
Heavy efforts have been put on developing new imaging system combining EVS and APS~\citeps{shariff2024event,lu2023learning,lu2023self}.
The introduction of EVS has nearly reshaped the entire framework of imaging formation and enhancement, impacting almost all relevant areas \eg, video super-resolution~\citeps{lu2023learning,jing2021turning}, video frame interpolation~\citeps{tulyakov2021time,tulyakov2022time,lu2023self}, deblurring~\citeps{yuan2007image, zhang2022deep, yunfan2023uniinr}, high dynamic range imaging~\citeps{xiaopeng2024hdr,messikommer2022multi}, low-light image enhancement~\citeps{wang2020experiment, liang2024towards}, and rolling shutter correction~\citeps{zhou2022evunroll,lu2023self}.
\textit{However, the majority of previous work focuses on using events as auxiliary information to boost the performance of classical RGB imaging systems, while methods and benchmarks that considering the challenges and opportunities of events in the APS ISP process, are lacking.}

\begin{figure}
\centering
\includegraphics[width=\linewidth]{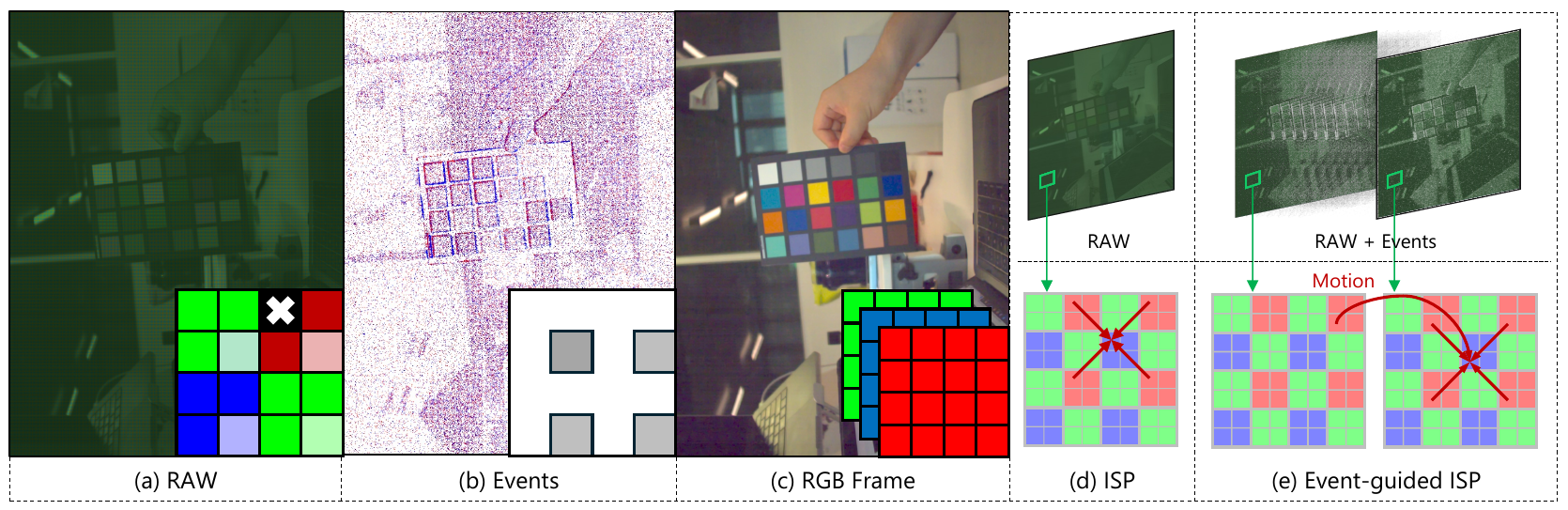}
\caption{\small (a), (b), and (c) display a RAW, Events, and RGB frame captured by the hybrid vision sensor (HVS), respectively. The RAW image follows a quad-Bayer pattern~\citeps{yang2022mipi}, while the events are positioned at the lower-right corner of each color pixel block, making the RAW resolution twice that of the events. (d) illustrates the traditional ISP process. (e) shows the potential event-guided ISP process, where the higher temporal resolution of events can captures motion information for ISP.}
\label{fig:0-CoverFigure-Release}
\end{figure}

Merging APS and EVS in ISP is non-trivial on the implementation level.
Prism spectrometer is an early stage attempt and it needs the corresponding optical mechanic setting~\citeps{tulyakov2022time}.
However, this prism-based approach is very cumbersome, requiring additional optical prisms and failing to ensure the alignment between APS and EVS.
Sensors that integrate both APS and EVS on the photodiode level are referred to as hybrid-vision sensors (HVS)~\citeps{hybridevs2024mipi3,mipi_2024}, which represent a cutting-edge technology, offering significant advancements in camera imaging.
Due to the manufacturing complexity and error-prone design process of HVS, the RAW data generated by APS in HVS exhibits higher noise, missing values at fixed positions, and is more sensitive to defects~\citeps{mipi_2024,hybridevs2024mipi3}.
Recent works have acknowledged this challenge and proposed datasets for demosaicing, denoising, or defect correction for APS RAW, where
\textit{the challenges in APS of HVS take precedence over the potential benefits events signal could provide.}
With the inherent higher dynamic range and lower latency, events can perceive a broader spectrum and capture more-instant motion information~\citeps{shekhar2022transform,liang2021cameranet}, allowing significant potential for boosting the denoising and color correction of ISP processing of APS RAW, as shown in Fig.~\ref{fig:0-CoverFigure-Release}.

To better explore the benefits of events on the ISP process of HVS, we propose a new dataset with \textbf{pixel-wise aligned} events and APS RAW image.
This dataset uses the under-development HVS-ALPIX-Eiger sensor~\citeps{alpsentek},
which rearranges event and APS in a quad-Bayer pattern (a quarter photodiodes are dedicated for event, as in Fig.\ref{fig:0-CoverFigure-Release}).
This sensor has a high resolution with $1224\times 1632$ for events and $2248\times 3264$ for RAW, and offers superior color and noise profiles compared to the DVS346~\citeps{scheerlinck2019ced}. These features make it promising for various applications~\citeps{lu2023learning}.
We ensure the dataset diversity in two ways: photographic setting and scenes.
For photographic setting, we adopt various values of aperture, focal length and exposure time.
For the scene diversity,  we cover 12 categories of scenes, across a wide range of color scenes, including flowers, buildings, under different weather and lighting conditions.
In total, 3373 APS frames and the corresponding events are captured.
A standard 24-color ColorChecker~\citeps{goto2003computer} is applied at certain frames as the color correction reference, as shown in Fig.~\ref{fig:0-CoverFigure-Release} (c).

To generate the ground truth RGB images for the dataset, we propose a controllable ISP framework based on MATLAB~\citeps{poon2001contemporary}.
This ISP framework, using the ColorChecker as a prior, performs tasks such as black level calculation~\citeps{li2010design}, demosaicing~\citeps{hirakawa2006joint}, white balance~\citeps{weng2005novel}, denoising~\citeps{abdelhamed2018high}, and color correction~\citeps{mcelvain2013camera}, resulting in high-quality RGB images with controllable errors as the reference ground truth.
Since the controllable framework requires the ColorChecker information as a prior, it cannot generalize to arbitrary scenes. The color accuracy and temporal stability of this ISP are also analyzed.
We categorize the existing ISP methods with RAW input into three categories and benchmark their performances on our dataset. We compare their performances across various scenarios and further conduct analysis on certain phenomena we have observed.
Additionally, we propose a simple UNet-like~\citeps{ronneberger2015u} event-guided ISP neural network to fuse events with RAW images.
This simple network can effectively improve the outdoor performance of ISP compared to the original UNet~\citeps{ronneberger2015u}.
We also identify key contributions and challenges of events in the ISP process, providing a foundation and direction for future research.

\section{Related Works}

\textbf{Event-guided Imaging Datasets:}
Event camera-guided imaging enhancement is an emerging field where the contribution of real datasets is crucial~\citeps{gallego2020event}.
Currently, event cameras have made significant progress in areas such as frame interpolation~\citeps{tulyakov2021time,lu2023self,niklaus2017video, bao2019depth}, video super-resolution~\citeps{lu2023learning,jing2021turning}, low-light enhancement~\citeps{liang2024towards,liang2023coherent}, and deblurring~\citeps{xu2021motion,lin2020learning,jiang2020learning}.
These advancements are supported by many foundational datasets~\citeps{tulyakov2021time,scheerlinck2019ced,lu2023learning}.
For example, BS-REGB~\citeps{tulyakov2022time} is a frame interpolation dataset using a beamsplitter to pair event cameras and RGB cameras.
The CED~\citeps{scheerlinck2019ced} dataset and APLEX-VSR~\citeps{lu2023learning} dataset have been used in research on event camera-guided video super-resolution.
Overall, these datasets serve as the cornerstone and pioneers in research on related tasks.
\textit{However, these datasets assume that event cameras can obtain high-quality RGB images through the ISP process, an assumption that is often too idealistic.}
Recognizing this, the MIPI~\citeps{hybridevs2024mipi3,mipi_2024} challenge introduced a RAW demosaic dataset for HVS in event cameras, addressing challenges like high noise and missing values in RAW from HVS.
\textit{Although this dataset is the first to focus on the RAW domain ISP process in event cameras, it lacks real event streams, thereby overlooking the potential role of events in the ISP process}.
To address this gap, we propose the \textbf{\textit{first}} dataset with aligned RAW and events  from a new HVS, aiming at exploring the potential value and role of event data in the ISP process.

\textbf{Learning-based ISP:}
Traditional ISPs~\citeps{schwartz2018deepisp} consist of long pipelines. In recent years deep learning has brought new insights to ISPs~\citeps{da2023isp} and has achieved higher performance.
These methods can be roughly categorized into three types.
The first type is full pipeline replacement methods, such as PyNet~\citeps{ignatov2020replacing} which use CNN architectures to replace the entire ISP pipeline.
The second type is stage-wise enhancement methods, like CameraNet~\citeps{liang2021cameranet} and AWNet~\citeps{dai2020awnet}, which divide the ISP pipeline into restoration and enhancement stages.
The third type is image enhancement network-based methods, which utilze state-of-the-art image proessing backbone models such as UNet~\citeps{ronneberger2015u} and Swin-Transformer~\citeps{liu2021swin} to deal with ISP tasks.
Though these methods have proven effective for RAW to RGB conversion, the potential of events in this process is not explored.

\textbf{Event-guided Image/Video Enhancement:}
Due to their high dynamic range and high temporal resolution~\citeps{gallego2020event,shariff2024event}, event cameras have garnered significant attention in the field of image/video enhancement and restoration~\citeps{gallego2020event,shariff2024event}, including many applications.
Initially, the use of events focused primarily on single-task enhancements of RGB images or videos ~\citeps{tulyakov2021time,pan2019bringing,lu2023learning}.
Recently, researchers recognized image enhancement tasks are inherently coupled with various degradations interwoven~\citeps{zhang2022unifying,song2022cir,yunfan2023uniinr}, suggesting a trend towards using events for unified solutions in camera computational imaging for multiple tasks.
\textit{However, existing methods focus \textbf{solely} on enhancing RGB images or videos using events, overlooking the ISP pipeline, which generate RGB images from RAW images. Additionally, existing methods neglect the potential value that events could provide in the ISP process.}

\section{Dataset Collection}
\begin{figure}[t!]
\includegraphics[width=\linewidth]{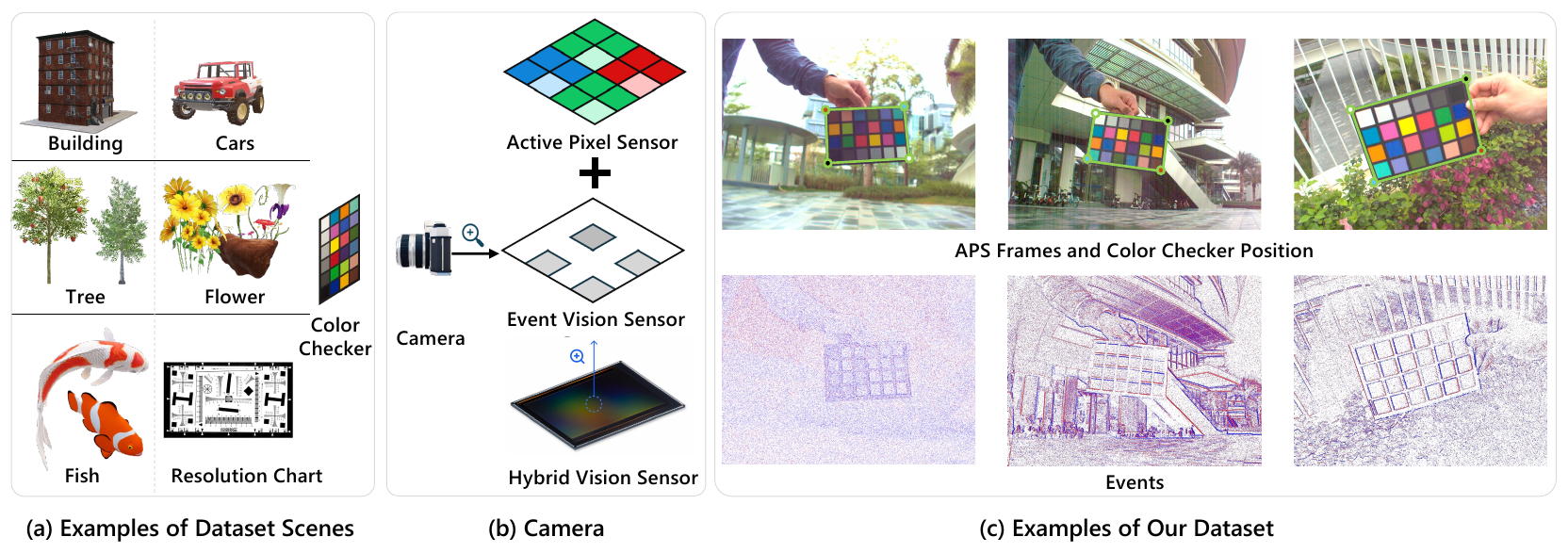}
\caption{{Overview of dataset collection. (a) illustrates the variety of scenes in the dataset, including buildings, plants, animals, and calibration boards. (b) presents a schematic of the HVS sensor, composed of a stacked active pixel sensor (APS) and an event vision sensor (EVS). (c) displays dataset samples.}\label{fig:2-DataCollection-Release}}
\end{figure}

As the first dataset, which we call HVS-ISP Dataset,  featuring paired raw-event data collected using a HVS, our aim is to facilitate research on event-guided RAW ISP.
We selected the HVS-Eiger sensor developed by ALPIX~\citeps{alpsentek}, which can output both APS and EVS signals that align in both time and space, as show in Fig.~\ref{fig:2-DataCollection-Release} (b).
More parameter details of APS and EVS are shown in Tab.~\ref{tab:comp_aps_evs}.
Compared to the Prophesee sensor~\citeps{tulyakov2021time}, which can only output event signals, and the DVS346 sensor~\citeps{scheerlinck2019ced}, which has lower resolution ($260\times 346$) and higher noise, our choice offers significant advantages.
Hence our dataset, captured with this advanced new sensor, holds significant value for the event vision research, providing a foundation resource for advanced exploration in event-guided RAW ISP.
The collection of the dataset focuses on two main aspects:
\textbf{(1)} the \textbf{diversity} of the dataset, ensuring it has broad representativeness to cover a wide range of real-world scenarios;
\textbf{(2)} the inclusion of a \textbf{ColorChecker} for ISP calibration, which helps the ISP accurately restore scene colors to generate high-quality RGB frames as references.

\begin{table}[t!]
\centering
\caption{Comparison between active pixel sensor (APS) and event vision sensor (EVS)~\citeps{alpsentek} in our dataset collection. APS and EVS are stacked together to form a hybrid-vision sensor (HVS).\label{tab:comp_aps_evs}}
\resizebox{1\linewidth}{!}{
\setlength{\tabcolsep}{0.039\linewidth}{
\begin{tabular}{cccccc}
\toprule
\textbf{Sensor}  &
\textbf{Resolution} &

\makecell[c]{\textbf{Frame}\\\textbf{Rate}} &
\makecell[c]{\textbf{Power}\\\textbf{Consumption}} &
\makecell[c]{\textbf{Redundant}\\ \textbf{Data Rate}} &
\makecell[c]{\textbf{Dynamic}\\ \textbf{Range}} \\
\midrule
\textbf{APS} & $2248\times 3264$ & 10$\sim$60 fps & $>$ 100 mW & 10 MB/s & 60 dB \\
\textbf{EVS} & $1124\times 1632$ & $\geq 800$ fps & $\sim$10 mW & 40-180 KB/s & $>$ 120 dB \\
\bottomrule
\end{tabular}
}}

\label{table:sensor_comparison}
\end{table}

\textbf{(1) Dataset Diversity:} In constructing our dataset, we paid particular attention to two types of diversity: camera parameter diversity and scene diversity.
\textbf{\textit{Camera Parameter Diversity:}} To ensure that our dataset encompasses a variety of photographic conditions, we made extensive adjustments to the camera parameters. This included aperture values ranging from $F1.0$ to $F6.0$, focal lengths extending from $8mm$ to $52mm$, and exposure times varying from $1 ms$ to $100 ms$.
\textbf{\textit{Scene Diversity:}} We focused on three key aspects to ensure comprehensive scene diversity:
\textit{Light Source Diversity:} We distinguished between indoor artificial light and outdoor natural light, with special consideration for different weather conditions. Data collection was performed under various lighting conditions, including sunny and cloudy days.
\textit{Motion Diversity:} We captured both dynamic and static videos, ensuring a mix of scenes with and without motion blur. This variety helps in testing and enhancing the performance of image processing algorithms under different motion conditions.
\textit{Material Diversity:} We included a wide array of scenes such as trees, plants, buildings, fish, dolls, and more. These scenes exhibit a broad spectrum of colors and textures, providing a rich dataset for comprehensive testing and improvement of image processing techniques.
\textbf{(2) ColorChecker as ISP Reference:}
To ensure precise color correction and white balance in ISP pipeline, we utilized a standard 24-color ColorChecker~\citeps{tian2002colour} as critical references.
At the start of each video shoot, we captured frames containing the ColorChecker and gradually removed the chart from subsequent frames.
We meticulously annotated the position of the ColorChecker in each frame using the LabelMe tool~\citeps{russell2008labelme}, as shown in Fig.~\ref{fig:2-DataCollection-Release} (c).
For frames without the ColorChecker, we applied previously determined ColorChecker parameters as references.
This approach guarantees reliable color correction data in our dataset.
Incorporating the ColorChecker allows generating high-quality RGB values, enhancing color fidelity.
This method ensures robustness for applications requiring accurate color restoration.
Additionally, we conducted a thorough manual review of the ColorChecker annotations to validate their accuracy, further improving our dataset's reliability for ISP algorithms.
In summary, based on these two main objectives, we captured a total of 24 videos.
Each video contains 80 to 140 frames, resulting in a total of 3373 APS RAW frames and their corresponding events.
Additionally, the dataset includes the positions of the ColorCheckers within the APS images.
We divided the dataset into training and test sets, with $3/4$ of the data used for training and $1/4$ for testing.
The testing set includes 3 indoor scenes and 3 outdoor scenes to ensure sufficient diversity.
\added{\textit{For more details on data collection and visualizations, please refer to the supplementary material.}}

\section{Controllable ISP}
\begin{figure}[t!]
\centering
\includegraphics[width=\linewidth]{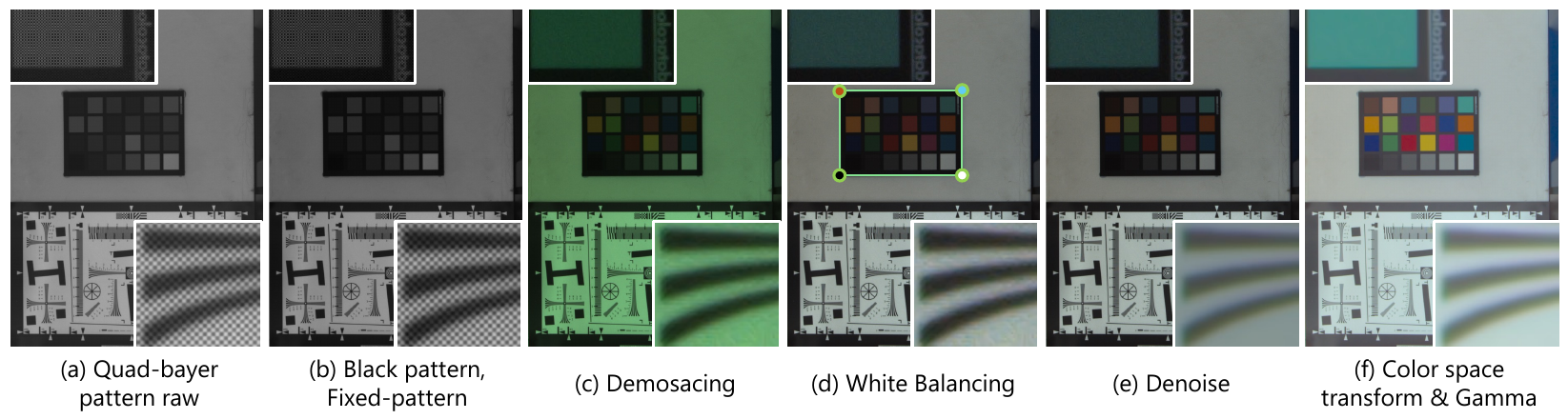}
\caption{\small Flows in controllable ISP process. (a) Quad-bayer pattern raw image, which serves as the initial input. (b) Black pattern and fixed-pattern noise removal to suppress sensor-induced artifacts. (c) Demosaicing to reconstruct a rgb image from the raw data. (d) White balancing using a ColorChecker for accurate color reproduction. (e) Denoising to filter out spatial noise from the image. (f) Color space transformation and Gamma to convert the image into the desired color space for final output.}
\label{fig:4-MATLAB_ISP-Release}
\end{figure}

\begin{wrapfigure}{lt}{240pt}
\centering
\includegraphics[width=\linewidth]{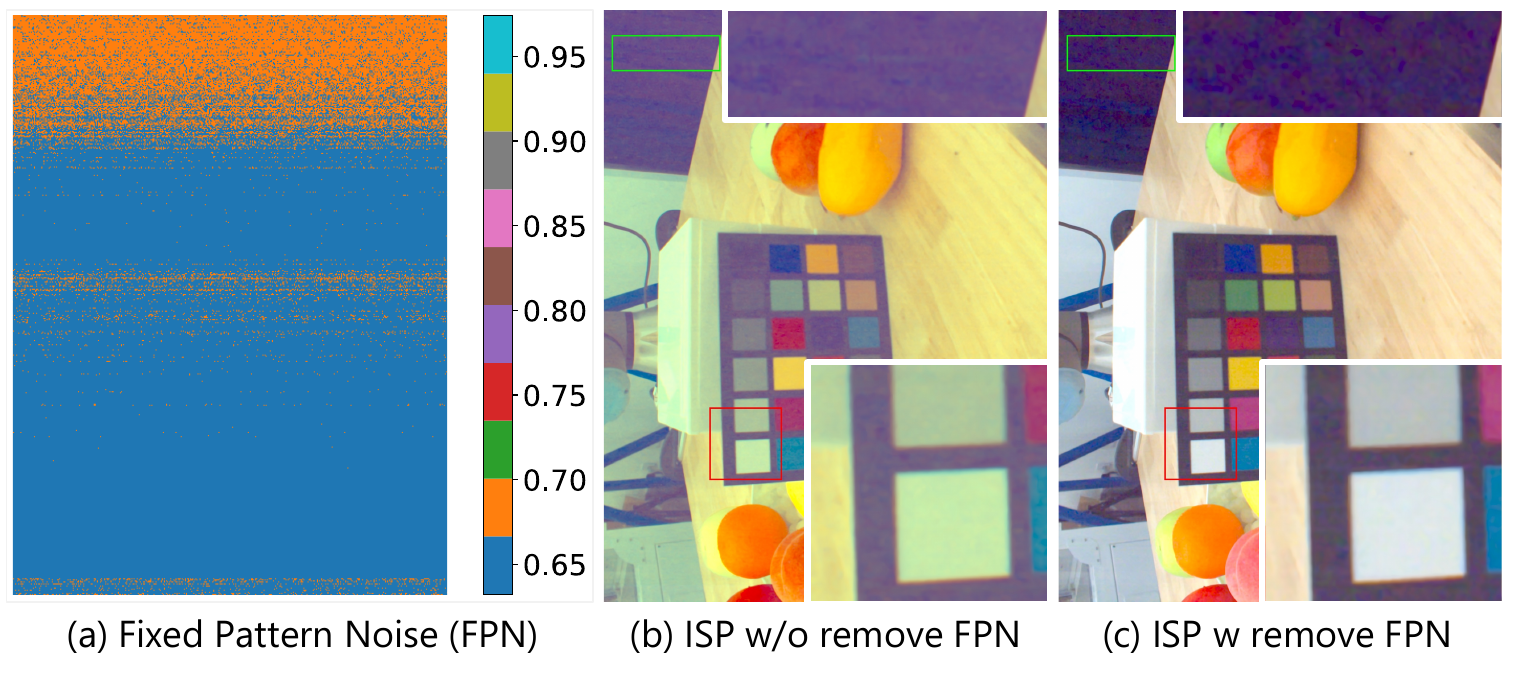}
\caption{\small Fixed pattern noise (FPN) removal. (a) Visualizes the camera's fixed pattern noise. (b) and (c) show the RGB images without and with fixed pattern noise removal, respectively. The image in (c) demonstrates lower noise and more accurate white balance after the removal of fixed pattern noise.}
\label{fig:5-FPN}
\end{wrapfigure}

The controllable ISP aims to provide module-based and analytically measurable RGB frames based on the APS RAW. With the support of the contained ColorChecker, the resulting frames have good color accuracy and low noise, serving as the reference for APS.
Requirement of the ColorChecker prevents from generalizing to other arbitrary scenes.
In this section, we introduce each module, followed by a quality evaluation and pros-and-cons discussion, with the hope that this ISP pipeline will be beneficial for the community.

\subsection{{Controllable ISP Pipeline}}
Fig.~\ref{fig:4-MATLAB_ISP-Release} depicts that how an image is processed via a conventional ISP pipeline, making the reference for the APS data.
\textbf{(1) Black Level and Fixed Pattern Subtraction:} Taking an arbitrary
 unprocessed bayer raw as input, a pre-calibrated global black level value \textit{blc} is subtracted, following by subtracting a fixed pattern vector \textit{fpn} \footnote{\textit{blc} and \textit{fpn} are calibrated in a pure-dark laboratory setting. Over five frames are captured and averaged to increase the calibration accuracy. }.  \textit{blc} is the min of a raw image taken under a pure-black environment while \textit{fpn} is a vector that records the per-row average value as the used sensor is only with horizontal fixed pattern, as shown in Fig.~\ref{fig:5-FPN}.
\textbf{(2) Demosaicing:}  Given bayer pattern, the well-adopted demosaicing method \citeps{quaddemosaic} is used. The resolution is preserved while the channel number is tripled. Note that this method is still prone to generating false color in very high frequency area, as shown in Fig.~\ref{fig:4-MATLAB_ISP-Release}(c).
\textbf{(3) Manual White Balancing:}  On a RGB image (greenish due to no white balance), we use LabelMe~\citeps{russell2008labelme} to extract the mean colors of 24 ColorChecker patches.
The 21$_{st}$ patch is used as the groundtruth illumination for manual white balance \cite{Qian_2017_ICCV, Qian_2019_CVPR}.
\textbf{(4) Spatial Denoising:} We use a milestone denoising method BM3D \citeps{dabov2009bm3d} to perform spatial denoising with the setting of $\sigma=50$.
\textbf{(5) Color Space transform:} Following Finlayson et.al. \citeps{finlayson2015color}, given the retrieved ColorChecker values and the predefined oracle ColorChecker values, we optimize towards the CIEDE00 error and obtain the final color correction matrix $\textit{ccm}$ of the shape $(3,3)$.  A linear sRGB image is then computed from the input image I: $I_{linsrgb} = I * \textit{ccm}$.
\textbf{(6) Gamma:} Following sRGB standard \citeps{anderson1996proposal}, a piecewise gamma curve is applied for brightness perception.
\added{\textit{Due to space limitations, please refer to the supplementary material for more details and hyperparameters of controllable ISP.}}

\subsection{{Controllable ISP Evaluation}}
\begin{figure}[t!]
    \centering
    \includegraphics[width=\linewidth]{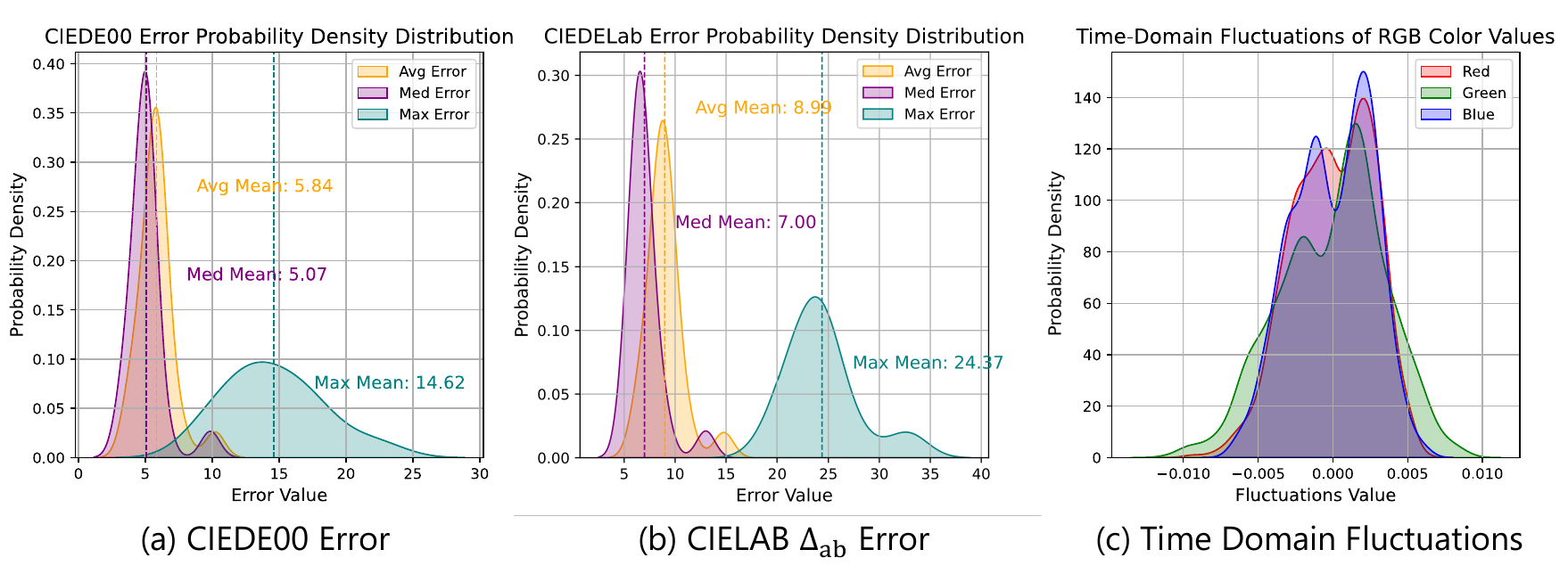}
    \vspace{-10pt}
    \caption{\small Color errors and fluctuations of our ISP method, computed using a ColorChecker. \textbf{(a)} CIEDE 2000 Error Probability Density Distribution: Displays CIEDE 2000 error values distribution with annotations for average (5.84), median (5.07), and maximum error means (14.62). \textbf{(b)} CIEDE Lab Error Probability Density Distribution: Shows CIEDE Lab error values distribution, indicating average (8.99), median (7.0), and maximum error means (24.37). \textbf{(c)} Time-Domain Fluctuations of RGB Color Values: Illustrates RGB color values fluctuations over time, representing temporal stability and variations in color accuracy.}
    \label{fig:10-ISP-ERROR-Release}
    \vspace{-10pt}
\end{figure}

We evaluated the controllable ISP in two main aspects: the \textbf{color accuracy} of individual images and the \textbf{temporal stability} of color recovery in continuous videos.
For color accuracy, we used the CIEDE00~\citeps{luo2001development} and CIELAB $\Delta_{ab}$~\citeps{lee2005comparison} metrics to evaluate color accuracy.
CIEDE00 is a widely used metric for color matching, considering the nonlinear characteristics of color differences and the human eye's sensitivity to colors, which accurately reflects human visual perception of color differences.
 CIELAB $\Delta_{ab}$ is a color difference metric based on the CIELAB color space~\citeps{mahy1994evaluation}.
Specifically, as shown in Fig.~\ref{fig:10-ISP-ERROR-Release} (a) (b), we conducted a ColorChecker-based evaluation on 100 randomly selected samples.
In CIEDE00~\citeps{luo2001development}, we obtained an average value of $5.84$ and a median value of $5.07$; For CIELAB $\Delta_{ab}$, we obtained an average value of $8.99$ and a median value of $7.00$, demonstrating that our method can generally restore colors up to an accurate level.
We displayed the maximum error distribution per image, showing that in CIEDE00 it is around $14$, and in CIELAB $\Delta_{ab}$ around $24$, affected by color filter sensitivity and photodiode layout.
For temporal stability in frame estimation differences, as shown in Fig.~\ref{fig:10-ISP-ERROR-Release} (c). We selected a 140-frame video, marking the ColorChecker in each frame. After generating colors frame by frame, we observed that differences for the 24 ColorChecker colors are all under $0.01$, mostly within $0.005$.
This confirms our algorithm's temporal stability.
In summary, we presented a controllable ISP pipeline and analyzed its performance.
However, the ISP contains numerous controllable variables and hyperparameters.
We hope that future researchers will focus on optimizing these controllable aspects of the ISP to further enhance its performance.

\section{Benchmark and Direction}

Based on the RGB frames obtained from the controllable ISP, we evaluate the performance of four types of ISP methods, particularly in outdoor and indoor scenarios.
The experiments are conducted in the same environment and framework.
Additionally, we will discuss the potential reasons behind these results and propose future research directions.
\textbf{Implementation Details:}
All our models were trained and tested on the same machine with a single A40 GPU with 48GB of GPU memory.
We used PyTorch~\citeps{paszke2017automatic} for all experiments, applying random cropping and rotation for data augmentation.
The training batch size was 1, with each patch sized at $1024 \times 1024$.
The learning rate was $0.0001$, and all models were trained for 50 epochs.
\textbf{Evaluation Metrics:} We evaluate model performances in two aspects: resource consumption, including parameters in millions ($M$), GFLOPS, and average inference time ($s$); and image reconstruction for indoor and outdoor scenes, measured by PSNR~\citeps{hore2010image}, SSIM~\citeps{brunet2011mathematical}, and $L_1$ distance.

\subsection{ISP Benchmark Methods}
Inspired by the prior ISP survey study~\citeps{da2023survey}, we categorize learning-based ISP models into three classes: full pipeline, stage-wise, image enhancement network-based. We selected two to four open-source models from each category for training and evaluation on our dataset. Furthermore, we put forward another new category of event fusion method, and since there is no prior research to refer to, we design a simple event-guided ISP neural network to test on our dataset.
\added{\textit{For more details on ISP methods, please refer to the supplementary material.}}

\textbf{Full Pipeline ISP:} These models utilize CNN architectures to integrate traditional ISP processes into an end-to-end conversion from RAW to RGB images.
Notable models in this category include PyNet~\citeps{ignatov2020replacing}, PyNetCA~\citeps{kim2020pynet}, InvertISP~\citeps{xing2021invertible}, and MV-ISPNet~\citeps{ignatov2020aim}.

\begin{table}[t]
\centering
\caption{\small Comparison on Parameters, FLOPS, and Time. Top two models are highlighted in \red{red} and \gre{green}.\label{table:method_comparison}}

\resizebox{1\linewidth}{!}{
\setlength{\tabcolsep}{0.007\linewidth}{
\begin{tabular}{l|cccccccc|cc}
\toprule
&
\multicolumn{1}{c}{\textbf{Unet}} &
\multicolumn{1}{c}{\textbf{PyNet}} &
\multicolumn{1}{c}{\textbf{CameraNet}} &
\multicolumn{1}{c}{\textbf{AWNet}} &
\multicolumn{1}{c}{\textbf{PyNetCA}} &
\multicolumn{1}{c}{\textbf{MW-ISPNet}} &
\multicolumn{1}{c}{\textbf{InvertISP}} &
\makecell[c]{\textbf{Swin} \\ \textbf{Transformer}} &

\textbf{eSL} &
\textbf{Ev-UNet} \\
\midrule
\textbf{Params}$\downarrow$     & {16.64} & 47.55 & 25.79 & 96.07 & 29.27 & \gre{7.22} & 92.44 & {8.87} & \red{0.737} & 21.51 \\
\hline
\textbf{GFLOPS}$\downarrow$      & \gre{4.52} & 111.96 & 19.19 & 120.21 & 51.27 & 29.22 & \red{1.41} & \blu{14.24} & 48.49 & 6.89 \\
\hline
\textbf{Time (s)}$\downarrow$   & \red{0.0100} & 0.0775 & {0.0300} & 0.2138 & \blu{0.0308} & 0.0459 & 0.0436 & 0.0868 & 0.063 & \gre{0.012} \\
\bottomrule
\end{tabular}
}}
\vspace{-10pt}
\end{table}

\begin{table}[t]
\centering
\caption{\small Comparison of Methods on
HVS ISP Dataset outdoor scenes. Top two models are highlighted in \red{red} and \gre{green}.
$^*$ refer to the results obtained by the same model with different hyperparameters.}

\resizebox{1\linewidth}{!}{
\setlength{\tabcolsep}{0.012\linewidth}{
\begin{tabular}{l ccc ccc ccc ccc}
\toprule
& \multicolumn{3}{c}{\texttt{2-Out-Tree-2}}
& \multicolumn{3}{c}{\texttt{3-Out-Flower-2}}
& \multicolumn{3}{c}{\texttt{4-Out-Building-1}}
& \multicolumn{3}{c}{Average} \\
& \textbf{PSNR}$\uparrow$   & \textbf{SSIM}$\uparrow$  & $L_1$$\downarrow$
& \textbf{PSNR}$\uparrow$   & \textbf{SSIM}$\uparrow$  & $L_1$$\downarrow$
& \textbf{PSNR}$\uparrow$   & \textbf{SSIM}$\uparrow$  & $L_1$$\downarrow$
& \textbf{PSNR}$\uparrow$   & \textbf{SSIM}$\uparrow$  & $L_1$$\downarrow$ \\
\midrule
\textbf{PyNET}
& \gre{31.70}   & \red{0.9818}  & \red{0.0190}  & \red{35.12}   & \red{0.9784}  & \red{0.0127}  & \red{30.60}& \red{0.9752} & \red{0.0223}  & \red{32.47}   & \red{0.9785}  & \red{0.0180}  \\
\textbf{PyNET$^*$}
& 27.56         & 0.9711        & 0.0310        & 32.35         & 0.9646        & 0.0175        & 28.20      & 0.9600       & 0.0311        & 29.37         & 0.9652        & 0.0265  \\
\textbf{PyNetCA}
& \red{31.86}   & \gre{0.9788}  & \gre{0.0202}  & \gre{34.19}   & \gre{0.9773}  & {0.0139}      & \gre{29.22}& \gre{0.9725} & 0.0280        & \gre{31.76}   & \gre{0.9762}  & \gre{0.0207}  \\
\textbf{InvertISP}
& 28.56         & 0.9487        & 0.0243        & 25.59         & 0.9298        & 0.0313        & 28.62      & 0.9307       & 0.0287        & 27.59         & 0.9364        & 0.0281  \\
\textbf{MV-ISPNet}
& 27.05         & 0.9680        & 0.0256        & 33.61         & 0.9648        & \gre{0.0137}  & 28.62      & 0.9657       & 0.0304        & 29.76         & 0.9662        & 0.0232  \\
\textbf{CameraNet}
& 11.18         & 0.2580        & 0.2289        & 12.39         & 0.2741        & 0.1899        & 10.52      & 0.2534       & 0.2609        & 11.36         & 0.2618        & 0.2266  \\
\textbf{CameraNet$^*$}
& 13.26         & 0.637        & 0.2044        & 13.59         & 0.2736        & 0.1770        & 10.06      & 0.2753       & 0.2474        & 12.30         & 0.3953        & 0.2096  \\
\textbf{AWNet}
& 14.33         & 0.8836        & 0.1166        & 20.10         & 0.9316        & 0.0519        & 16.70      & 0.9390       & 0.0951        & 17.04         & 0.9180        & 0.0879  \\
\textbf{Swin-Transformer}
& 25.02         & 0.9539        & 0.0308        & 29.14         & 0.9555        & 0.0231        & 21.57      & 0.9295       & 0.0523        & 25.24         & 0.9463        & 0.0354  \\
\textbf{UNet}
& 21.97         & 0.9583        & 0.0393        & 29.43         & 0.9717        & 0.0208        & 22.12      & 0.9603       & 0.0460        & 24.51         & 0.9634        & 0.0354  \\
\textbf{UNet$^*$}
& 29.52         & 0.9752        & {0.0206}      & 25.75         & 0.9623        & 0.0323        & 29.24      & 0.9680       & \gre{0.0265}  & 28.17         & 0.9685        & 0.0265  \\
\hline
\textbf{eSL-Net}
& 25.67 	    & 0.9424 	    & 0.0342 	    & 19.39 	    &0.9180 	    & 0.0576 	    & 24.01 	 & 0.9277 	    & 0.0502 	    & 23.02 	    & 0.9294 	    & 0.0473  \\
\textbf{EV-UNet}~~                  & 32.86 	    & 0.9795 	    & 0.0148 	    & 32.87 	    &0.9698 	    & 0.0157 	    & 24.59 	 & 0.9600 	    & 0.0369 	    & 30.11 	    & 0.9698 	    & 0.0225 \\

\bottomrule
\label{table:method_outdoor_performance}
\end{tabular}
}}
\vspace{-10pt}
\end{table}

\begin{figure}[t!]
\centering
\includegraphics[width=\linewidth]{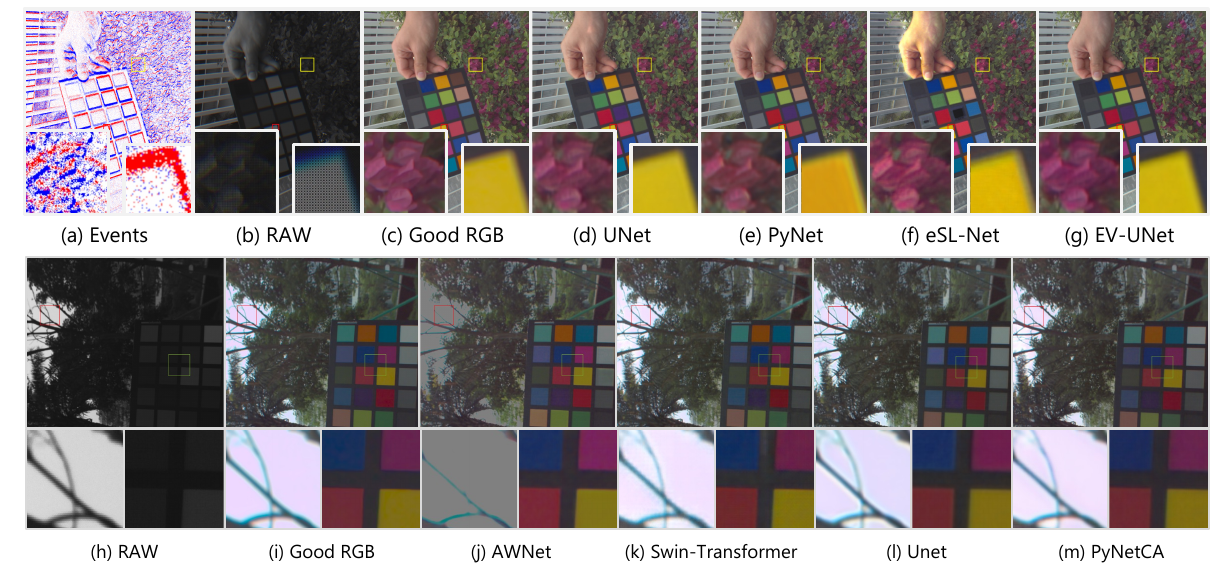}
\vspace{-10pt}
\caption{\small \added{Visualization results of different methods on HVS-ISP Dataset outdoor scenes.}}
\label{fig:8-Compre-Vis-Outdoor-with-events}
\label{fig:8-Compre-Vis-Outdoor}
\vspace{-10pt}
\end{figure}

\textbf{Stage-wise ISP:} They employ multiple specialized modules to handle different ISP tasks, either sequentially or in parallel, to produce the final image. In our benchmark, we selected CameraNet~\citeps{liang2021cameranet} and AWNet~\citeps{dai2020awnet} for their distinct approaches.
Note that due to the unavailability of a PyTorch version of CameraNet~\citeps{liang2021cameranet}, we experimented on a converted version. The modules in the original AWNet~\citeps{dai2020awnet} are trained independently, however in our experiment we trained them end-to-end.

\textbf{Image Enhancement Network-Based ISP:} There have been numbers of high performance backbone models for image enhancement in image enhancement tasks like deblurring~\citeps{zhang2022deep} and super-resolution~\citeps{chen2022real}.
Though not initially designed for ISPs, minor modifications can adapt these models for ISP tasks. For our benchmark, we selected UNet~\citeps{ronneberger2015u} and Swin-Transformer~\citeps{liu2021swin,lu2024event}.

\textbf{Event Fusion Method:}
As the first research on event-guided ISP, we have no prior research for reference.
Therefore, we selected eSL-Net~\citeps{wang2020event}, an event-based backbone network used in various tasks~\citeps{lu2023learning}.
Additionally, we merged events as voxel-grid~\citeps{liu2023voxel} with UNet's encoder as EV-UNet to verify events effectiveness and challenges.

\begin{table}[t!]
\centering
\caption{Comparison of Methods on HVS ISP Dataset indoor scenes. $^*$ refer to the results obtained by the same model with different hyperparameters.}

\resizebox{1\linewidth}{!}{
\setlength{\tabcolsep}{0.012\linewidth}{
\begin{tabular}{lcccccccccccc}
\toprule
&
\multicolumn{3}{c}{\texttt{1-In-Fruit-2}}
&
\multicolumn{3}{c}{\texttt{3-In-ColChecker-40}}
&
\multicolumn{3}{c}{\texttt{4-In-RLChart-10}}
&
\multicolumn{3}{c}{\texttt{Average}} \\
\textbf{Methods}
& \textbf{PSNR}$\uparrow$   & \textbf{SSIM}$\uparrow$  & $L_1$$\downarrow$
& \textbf{PSNR}$\uparrow$   & \textbf{SSIM}$\uparrow$  & $L_1$$\downarrow$
& \textbf{PSNR}$\uparrow$   & \textbf{SSIM}$\uparrow$  & $L_1$$\downarrow$
& \textbf{PSNR}$\uparrow$   & \textbf{SSIM}$\uparrow$  & $L_1$$\downarrow$ \\
\midrule
\textbf{PyNET}
& 13.09           & 0.7970        & 0.2182        & 11.38       & 0.7922       & 0.2489       & 11.42        & 0.7100        & 0.2563       & 11.97       & 0.7664        & 0.2412  \\
\textbf{PyNET$^*$}
& 14.46           & 0.8068        & 0.2008        & 24.02       & 0.9550       & 0.0497       & 13.58        & 0.7694        & 0.1978       & 17.36       & 0.8437        & 0.1494  \\
\textbf{PyNetCA}
& 18.13           & 0.8843        & 0.1253        & 29.53       & 0.9723       & 0.0246       & \red{35.51}  & \red{0.9727}  & \red{0.0121} & 27.72       & 0.9431        & 0.0540  \\
\textbf{InvertISP}
& 25.83           & 0.9098        & 0.0346        & 28.33       & 0.9500       & \gre{0.0235} & 30.65        & 0.9578        & 0.0183       & 28.27       & 0.9392        & 0.0254  \\
\textbf{MV-ISPNet}
& \gre{31.91}     & \gre{0.9594}  & \gre{0.0185}  & \gre{29.56} & \gre{0.9729} & 0.0265       & 31.88        & 0.9670        & 0.0170       & \gre{31.12} & \gre{0.9664}  & \gre{0.0207}  \\
\textbf{CameraNet}
& 13.06           & 0.2660        & 0.1947        & 13.58       & 0.2722       & 0.1836       & 12.47        & 0.2391        & 0.2257       & 13.04       & 0.2591        & 0.2013  \\
\textbf{CameraNet$^*$}
& 14.18           & 0.290        & 0.1630        & 10.60       & 0.2667       & 0.2545       & 13.26        & 0.2636        & 0.2044       & 12.68       & 0.2672        & 0.2073  \\
\textbf{AWNet}
& 17.95           & 0.8665        & 0.1302        & \red{32.17} & \red{0.9807} & \red{0.0184} & 30.98        & 0.9596        & 0.0215       & 27.03       & 0.9356        & 0.0567  \\
\textbf{Swin-Transformer}
& 25.73           & 0.9397        & 0.0301        & 25.50       & 0.9561       & 0.0359       & 26.18        & 0.9486        & 0.0252       & 25.80       & 0.9481        & 0.0304  \\
\textbf{UNet}
& 17.62           & 0.9161        & 0.0747        & 13.96       & 0.8828       & 0.1454       & 15.53        & 0.8750        & 0.1170       & 15.70       & 0.8913        & 0.1124  \\
\textbf{UNet$^*$}
& \red{32.52}     & \red{0.9659}  & \red{0.0161}  & 29.04       & 0.9740       & 0.0257       & \gre{33.72}  & \gre{0.9716}  & \gre{0.0146} & \red{31.76} & \red{0.9705}  & \red{0.0188}  \\
\hline
\textbf{eSL}
& 27.09 	&0.9428 	&0.0331 &	24.79 &	0.9548 &	0.0434& 	26.52 &	0.9415 &	0.0379 	&26.13 &	0.9464 	&0.0381  \\
\textbf{EV-UNet}          & 14.16 	&0.8706 &	0.1533 	&31.64 &	0.9779 &	0.0214 &	32.33 &	0.9678 &	0.0173 &	26.04 &	0.9388 &	0.0640 \\

\toprule
\label{table:method_indoor_performance}
\end{tabular}
}}
\vspace{-10pt}
\end{table}

\subsection{Comparative Experiments and Visualization Analysis}
\textbf{Computational Performance:}
In Tab.~\ref{table:method_comparison}, InvertISP~\citeps{xing2021invertible} excels in computational efficiency with $1.41$ GFLOPS, significantly lower than the over $100$ GFLOPS of AWNet~\citeps{dai2020awnet} and PyNet~\citeps{kim2020pynet}, which is suitable for limited computing resources. UNet surpasses CameraNet~\citeps{liang2021cameranet} in processing speed with a response time of 0.01 s, preferable for real-time performance. Overall, UNet demonstrates balanced performance with low GFLOPS and the fastest processing speed, due to its straightforward design.

\textbf{Outdoor Performance:}
Tab.~\ref{table:method_outdoor_performance} shows the superior performance of PyNet across three outdoor backgrounds.
PyNet~\citeps{kim2020pynet} achieves the best PSNR~\citeps{hore2010image}, SSIM~\citeps{brunet2011mathematical}, and $L_1$ with an overall average PSNR~\citeps{hore2010image} of $32.47$, significantly higher than other models.
Specifically, EV-UNet shows {significant improvement in outdoor scenes} with UNet after incorporating events gain, increasing from $28.17$ to $30.11$.
In contrast, the commonly used event-based method eSL-Net performs poorly with a PSNR~\citeps{hore2010image} of only $23$.
This poor performance mainly results from the \textbf{\textit{limited receptive field}} of eSL, which is insufficient for estimating the \textbf{\textit{global illumination}} information, and thus failing to achieve consistent global illumination enhancement.
We further discuss on this issue in Sec.~\ref{sec:discusstion}.
we also visualize the results in Fig.~\ref{fig:8-Compre-Vis-Outdoor-with-events}. PyNet has achieved the highest PSNR~\citeps{hore2010image} but exhibits edge artifacts, this is likely due to the overfitting of the model.
In outdoor scenes, event-enhanced outputs of EV-UNet show good global consistency. Fig.~\ref{fig:8-Compre-Vis-Outdoor} shows that AWNet~\citeps{dai2020awnet} struggles with fine texture restoration, explaining its inferior performance to other methods.

\begin{figure}[t!]
    \centering
    \includegraphics[width=\linewidth]{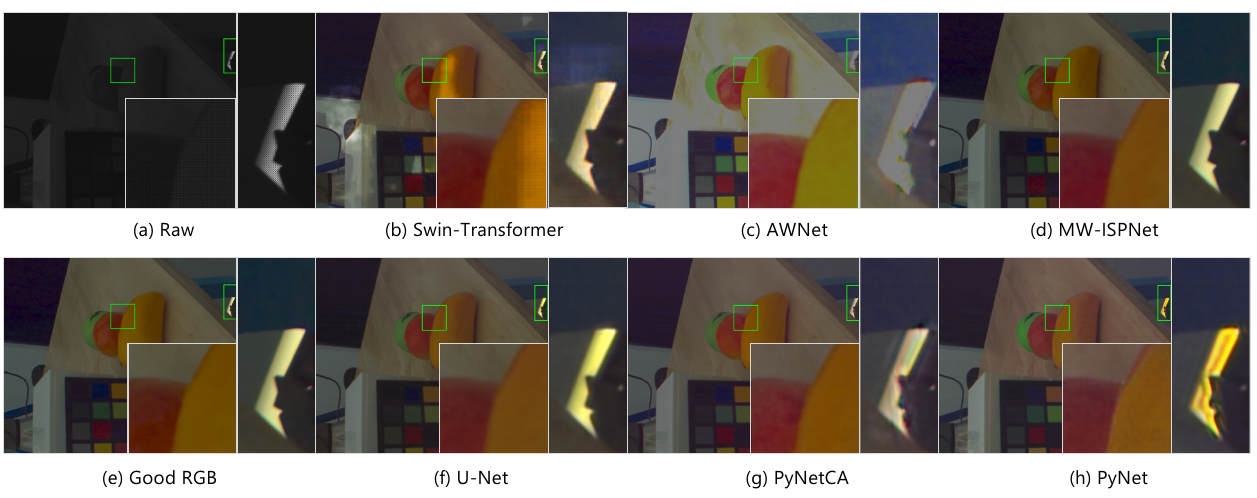}
    \vspace{-15pt}
    \caption{\small \added{Visualizations on HVS-ISP Dataset indoor scenes.}}
    \label{fig:9-Compre-Vis-Indoor}
\vspace{-10pt}
\end{figure}

\textbf{Indoor Performance:}
Tab.~\ref{table:method_indoor_performance} shows that UNet$^*$ excels in indoor environments, especially when handling multiple colored fruits and scenes with complex lighting and details.
The output of AWNet~\citeps{dai2020awnet} has overall excessive brightness, as illustrated in Fig.~\ref{fig:9-Compre-Vis-Indoor}, explaining its low PSNR values.
PyNet exhibits noticeable artifacts, consistent with the good RGB edge but with significantly different brightness, likely due to the ill-posed nature of brightness recovery in the ISP process, resulting in its poor indoor performance.
Event-fusion methods perform poorly indoors, primarily due to flickering light sources that complicate event characteristics. For more analysis about these issues, please refer to Sec.~\ref{sec:discusstion}.

\textbf{Summary:}
These sections show that the performance of numerous ISP methods on HVS sensor datasets varies significantly across different scenes. For instance, PyNet and AWNet~\citeps{dai2020awnet} exhibit great variability between indoor and outdoor environments, underscoring that learning-based ISP methods are highly scene-dependent. This highlights the necessity for future work to analyze different scenes individually to fully understand the performance of a network.
Furthermore, adding events to UNet significantly improves performance in outdoor scenarios but not indoors, mainly due to the flickering indoor lighting. Addressing this issue remains a crucial challenge for future research.

\subsection{Discussion and Future Direction\label{sec:discusstion}}
Through the comprehensive and objective evaluation of various models on our dataset, we have also observed a number of findings that can bring insights for future work.

\textbf{Significant Indoor-Outdoor Performance Gap on PyNet and AWNet~\citeps{dai2020awnet}:}
We observed a significant indoor-outdoor performance gap on PyNet~\citeps{ignatov2020replacing} and AWNet~\citeps{dai2020awnet}. PyNet performs better in outdoor scenes than indoor, ranking the top of all models, while AWNet~\citeps{dai2020awnet} shows quite the opposite behavior. Generally, outdoor scenes have more dynamic and varied lighting compared to indoor environments, which are difficult for models to learn. The original AWNet~\citeps{dai2020awnet} is designed to be trained in a multi-stage manner with different loss functions. Therefore it might have fallen into sub-optima when trained end-to-end in our experiment, resulting in the poor performance in modeling the harder outdoor scenes.

\begin{wrapfigure}{lt}{250pt}
    \centering
    \includegraphics[width=\linewidth]{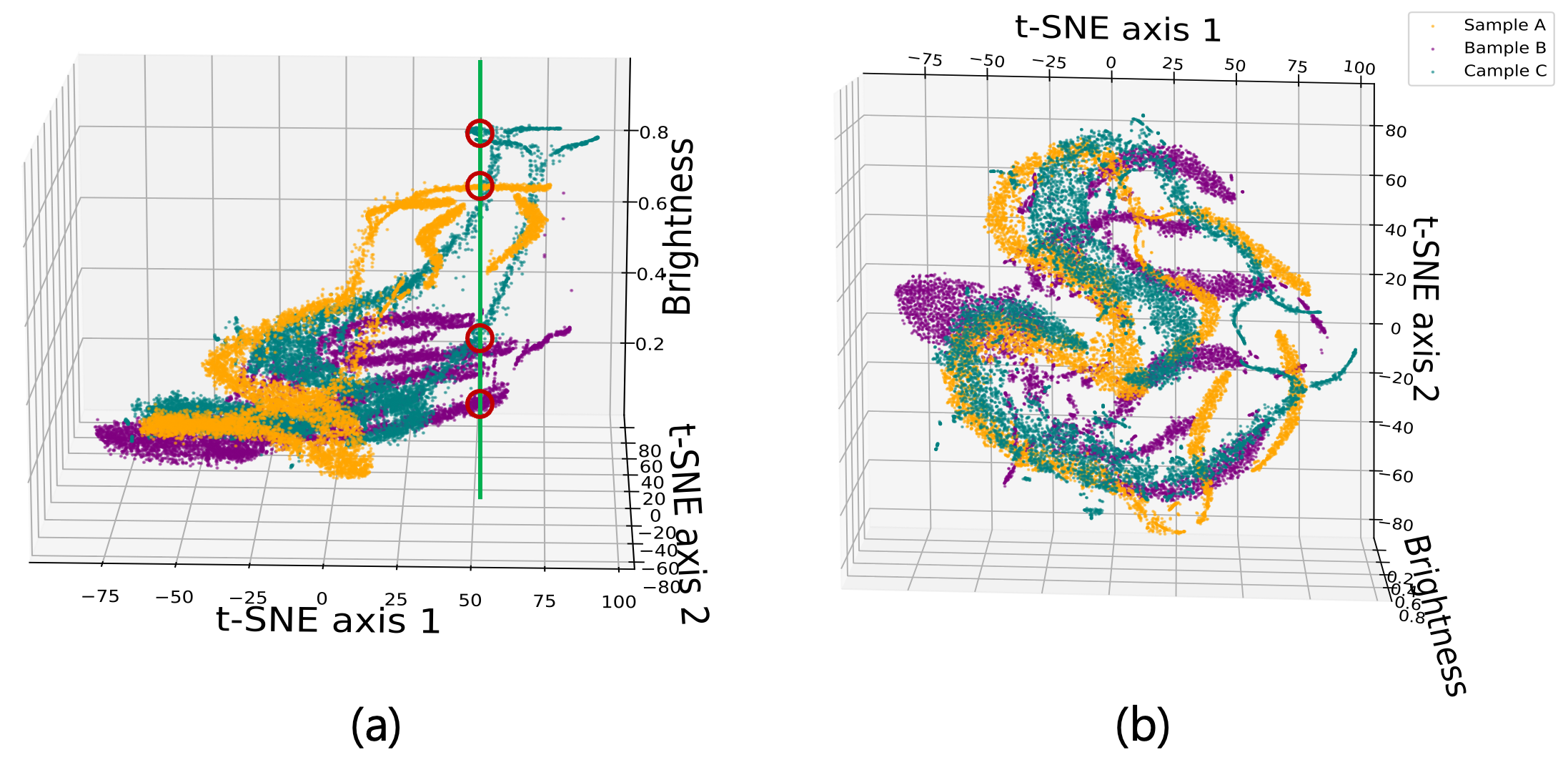}
    \caption{\small The ill-posedness of brightness estimation in the ISP process. We visualized the $5 \times 5$ region in the RAW image and the brightness of corresponding pixel in the color image at the center of this region.
    The results show that the same RAW region corresponds to different brightness levels in different images.}
    \label{fig:12-RAW-Y-TSEN}
\end{wrapfigure}

\textbf{Local Brightness Artifacts:}
Artifacts occur when the brightness in certain image areas significantly deviates from the overall luminance (see Fig.~\ref{fig:9-Compre-Vis-Indoor}). We investigated this by examining the relationship between a brightness of a pixel and the RAW data within its $5 \times 5$ vicinity. We treat the neighboring RAW data as a 25-dimensional vector, and apply t-SNE to project it onto a 2D plane, recording the $(x,y)$ coordinates. We then converted the RGB values of the pixel to YUV, recording the Y (brightness) as the $z$ coordinate, as shown in Fig.~\ref{fig:12-RAW-Y-TSEN}.
By plotting pixels from three random images in 3D (Fig.~\ref{fig:12-RAW-Y-TSEN}), we show that pixel brightness and neighboring RAW data have a non-injective relationship.
Multiple brightness levels can emerge from the same RAW data, indicating that \textbf{global information}, not just local RAW value, is essential for accurately determining pixel brightness to avoid local artifacts.

\textbf{Event Gains:} The integration of events in our dataset significantly enhances performance in outdoor scenes when comparing EV-UNET with UNet, primarily due to the additional motion information and dynamic range provided by the events.
However, simple fusion does not fully exploit these characteristics, highlighting the need for more sophisticated designs in future research.
Conversely, performance decreases in indoor scenes, primarily due to the flickering of artificial light sources.

\textbf{Flickering Artificial Lighting:}
Under certain indoor scenarios, some artificial light source ~\citeps{xu2023seeing}, \eg LEDs, flicker because of the alternating current frequency. Given that the event frame rate of the sensor significantly exceeds the usual AC frequency ($50$ or $60$ Hz), the flickering lighting introduces considerable fluctuations in the event data over time. The distributions and features of events in these conditions are completely different from that in the natural lighting conditions, and could result in the model's failure in restoring the images from RAW data.

\section{Conclusion}
In this work, we present the first events-RAW paired dataset for event-guided ISP research.
The dataset consists of $3373$ high quality high resolution RAW images and corresponding \textbf{pixel-level aligned} events.
Subsequently, good RGB frames are generated by a controllable ISP pipeline we proposed.
A comprehensive evaluation and analysis of existing learnable ISPs and a simple event-guided ISP method are conducted on our dataset.
Based on this analysis, we summarize some key points and challenges for event-guided ISP.
We wish to emphasize the potential of event data in ISP processes again. Event cameras have a high dynamic range and high temporal resolution, which surpass the limits of human vision systems. In terms of dynamic range and temporal sampling, the information captured by event sensor is somehow a superset of that of human eye. Therefore, generating images perceptible to human vision is a matter of downward compatibility.
\textbf{Limitations:}
Firstly, the scale of our dataset is relatively small, because the HVS sensor we use is still in the prototype stage and the associated hardware is cumbersome and exhibits low stability, which has raised the cost in data collection and thus a limited size dataset. And yet we are committed to expanding the dataset with more diverse real-world scenarios in future research.
Secondly, our dataset has not thoroughly addressed the issue of flickering in artificial lighting caused by alternating current, especially in indoor scenarios.
The flickering considerably impairs the performance of our method and further research should pay attention to this problem.

\textbf{Acknowledgements:}
This work was supported in part by the National Key R\&D Program of China (Grant No.2023YFF0725001),in part by the National Natural Science Foundation of China (Grant No.92370204), in part by the Guangdong Basic and Applied Basic Research Foundation (Grant No.2023B1515120057), in part by Guangzhou-HKUST(GZ) Joint Funding Program (Grant No.2023A03J0008), Education Bureau of Guangzhou Municipality.

{\small
\bibliography{egbib}

\begin{thebibliography}{73}
\providecommand{\natexlab}[1]{#1}
\providecommand{\url}[1]{\texttt{#1}}
\expandafter\ifx\csname urlstyle\endcsname\relax
  \providecommand{\doi}[1]{doi: #1}\else
  \providecommand{\doi}{doi: \begingroup \urlstyle{rm}\Url}\fi

\bibitem[Abdelhamed et~al.(2018)Abdelhamed, Lin, and Brown]{abdelhamed2018high}
Abdelrahman Abdelhamed, Stephen Lin, and Michael~S Brown.
\newblock A high-quality denoising dataset for smartphone cameras.
\newblock In \emph{Proceedings of the IEEE conference on computer vision and
  pattern recognition}, pp.\  1692--1700, 2018.

\bibitem[{Alpsentek}(2024)]{alpsentek}
{Alpsentek}.
\newblock Alpix-eiger product overview:~\url{https://alpsentek.com/product},
  2024.
\newblock URL \url{https://alpsentek.com/product}.
\newblock Accessed: 2024-05-19.

\bibitem[Anderson et~al.(1996)Anderson, Motta, Chandrasekar, and
  Stokes]{anderson1996proposal}
Matthew Anderson, Ricardo Motta, Srinivasan Chandrasekar, and Michael Stokes.
\newblock Proposal for a standard default color space for the internet—srgb.
\newblock In \emph{Color and imaging conference}, volume~4, pp.\  238--245.
  Society of Imaging Science and Technology, 1996.

\bibitem[Bao et~al.(2019)Bao, Lai, Ma, Zhang, Gao, and Yang]{bao2019depth}
Wenbo Bao, Wei-Sheng Lai, Chao Ma, Xiaoyun Zhang, Zhiyong Gao, and Ming-Hsuan
  Yang.
\newblock Depth-aware video frame interpolation.
\newblock In \emph{Proceedings of the IEEE/CVF conference on computer vision
  and pattern recognition}, pp.\  3703--3712, 2019.

\bibitem[Brunet et~al.(2011)Brunet, Vrscay, and Wang]{brunet2011mathematical}
Dominique Brunet, Edward~R Vrscay, and Zhou Wang.
\newblock On the mathematical properties of the structural similarity index.
\newblock \emph{IEEE Transactions on Image Processing}, 21\penalty0
  (4):\penalty0 1488--1499, 2011.

\bibitem[Chen et~al.(2022)Chen, He, Qing, Wu, Ren, Sheriff, and
  Zhu]{chen2022real}
Honggang Chen, Xiaohai He, Linbo Qing, Yuanyuan Wu, Chao Ren, Ray~E Sheriff,
  and Ce~Zhu.
\newblock Real-world single image super-resolution: A brief review.
\newblock \emph{Information Fusion}, 79:\penalty0 124--145, 2022.

\bibitem[da~Silva et~al.(2023{\natexlab{a}})da~Silva, da~Silva, Arrais,
  de~Ara{\'u}jo~Neto, Lopes, Bileki, Lima, Rondon, de~Souza, Regazio,
  et~al.]{da2023isp}
Matheus Henrique~Marques da~Silva, Jhessica Victoria~Santos da~Silva,
  Rodrigo~Reis Arrais, Wladimir Barroso~Guedes de~Ara{\'u}jo~Neto,
  Leonardo~Tadeu Lopes, Guilherme~Augusto Bileki, Iago~Oliveira Lima,
  Lucas~Borges Rondon, Bruno~Melo de~Souza, Mayara~Costa Regazio, et~al.
\newblock Isp meets deep learning: A survey on deep learning methods for image
  signal processing.
\newblock \emph{arXiv preprint arXiv:2305.11994}, 2023{\natexlab{a}}.

\bibitem[da~Silva et~al.(2023{\natexlab{b}})da~Silva, da~Silva, Arrais, Neto,
  Lopes, Bileki, Lima, Rondon, de~Souza, Regazio, et~al.]{da2023survey}
Matheus Henrique~Marques da~Silva, Jhessica Victoria~Santos da~Silva,
  Rodrigo~Reis Arrais, Wladimir Barroso Guedes de~Ara{\'u}jo Neto,
  Leonardo~Tadeu Lopes, Guilherme~Augusto Bileki, Iago~Oliveira Lima,
  Lucas~Borges Rondon, Bruno~Melo de~Souza, Mayara~Costa Regazio, et~al.
\newblock Survey on software isp methods based on deep learning.
\newblock \emph{arXiv preprint arXiv:2305.11994}, 2023{\natexlab{b}}.

\bibitem[Dabov et~al.(2009)Dabov, Foi, Katkovnik, and
  Egiazarian]{dabov2009bm3d}
Kostadin Dabov, Alessandro Foi, Vladimir Katkovnik, and Karen Egiazarian.
\newblock Bm3d image denoising with shape-adaptive principal component
  analysis.
\newblock In \emph{SPARS'09-Signal Processing with Adaptive Sparse Structured
  Representations}, 2009.

\bibitem[Dai et~al.(2020)Dai, Liu, Li, and Chen]{dai2020awnet}
Linhui Dai, Xiaohong Liu, Chengqi Li, and Jun Chen.
\newblock Awnet: Attentive wavelet network for image isp.
\newblock In \emph{Computer Vision--ECCV 2020 Workshops: Glasgow, UK, August
  23--28, 2020, Proceedings, Part III 16}, pp.\  185--201. Springer, 2020.

\bibitem[Delbracio et~al.(2021)Delbracio, Kelly, Brown, and
  Milanfar]{delbracio2021mobile}
Mauricio Delbracio, Damien Kelly, Michael~S Brown, and Peyman Milanfar.
\newblock Mobile computational photography: A tour.
\newblock \emph{Annual review of vision science}, 7:\penalty0 571--604, 2021.

\bibitem[Finlayson et~al.(2015)Finlayson, Mackiewicz, and
  Hurlbert]{finlayson2015color}
Graham~D Finlayson, Michal Mackiewicz, and Anya Hurlbert.
\newblock Color correction using root-polynomial regression.
\newblock \emph{IEEE Transactions on Image Processing}, 24\penalty0
  (5):\penalty0 1460--1470, 2015.

\bibitem[Gallego et~al.(2020)Gallego, Delbr{\"u}ck, Orchard, Bartolozzi, Taba,
  Censi, Leutenegger, Davison, Conradt, Daniilidis, et~al.]{gallego2020event}
Guillermo Gallego, Tobi Delbr{\"u}ck, Garrick Orchard, Chiara Bartolozzi, Brian
  Taba, Andrea Censi, Stefan Leutenegger, Andrew~J Davison, J{\"o}rg Conradt,
  Kostas Daniilidis, et~al.
\newblock Event-based vision: A survey.
\newblock \emph{IEEE transactions on pattern analysis and machine
  intelligence}, 44\penalty0 (1):\penalty0 154--180, 2020.

\bibitem[Goto et~al.(2003)Goto, Dogru, Kojima, and Tsubota]{goto2003computer}
Eiki Goto, Murat Dogru, Takashi Kojima, and Kazuo Tsubota.
\newblock Computer-synthesis of an interference color chart of human tear lipid
  layer, by a colorimetric approach.
\newblock \emph{Investigative ophthalmology \& visual science}, 44\penalty0
  (11):\penalty0 4693--4697, 2003.

\bibitem[Hirakawa \& Parks(2006)Hirakawa and Parks]{hirakawa2006joint}
Keigo Hirakawa and Thomas~W Parks.
\newblock Joint demosaicing and denoising.
\newblock \emph{IEEE Transactions on Image Processing}, 15\penalty0
  (8):\penalty0 2146--2157, 2006.

\bibitem[Hore \& Ziou(2010)Hore and Ziou]{hore2010image}
Alain Hore and Djemel Ziou.
\newblock Image quality metrics: Psnr vs. ssim.
\newblock In \emph{2010 20th international conference on pattern recognition},
  pp.\  2366--2369. IEEE, 2010.

\bibitem[Huang et~al.(2017)Huang, Chen, Ceylan, and Jin]{huang20176}
Jingwei Huang, Zhili Chen, Duygu Ceylan, and Hailin Jin.
\newblock 6-dof vr videos with a single 360-camera.
\newblock In \emph{2017 IEEE Virtual Reality (VR)}, pp.\  37--44. IEEE, 2017.

\bibitem[Ignatov et~al.(2020{\natexlab{a}})Ignatov, Timofte, Zhang, Liu, Wang,
  Zuo, Zhang, Zhang, Peng, Ren, et~al.]{ignatov2020aim}
Andrey Ignatov, Radu Timofte, Zhilu Zhang, Ming Liu, Haolin Wang, Wangmeng Zuo,
  Jiawei Zhang, Ruimao Zhang, Zhanglin Peng, Sijie Ren, et~al.
\newblock Aim 2020 challenge on learned image signal processing pipeline.
\newblock In \emph{Computer Vision--ECCV 2020 Workshops: Glasgow, UK, August
  23--28, 2020, Proceedings, Part III 16}, pp.\  152--170. Springer,
  2020{\natexlab{a}}.

\bibitem[Ignatov et~al.(2020{\natexlab{b}})Ignatov, Van~Gool, and
  Timofte]{ignatov2020replacing}
Andrey Ignatov, Luc Van~Gool, and Radu Timofte.
\newblock Replacing mobile camera isp with a single deep learning model.
\newblock In \emph{Proceedings of the IEEE/CVF conference on computer vision
  and pattern recognition workshops}, pp.\  536--537, 2020{\natexlab{b}}.

\bibitem[Ingle \& Phute(2016)Ingle and Phute]{ingle2016tesla}
Shantanu Ingle and Madhuri Phute.
\newblock Tesla autopilot: semi autonomous driving, an uptick for future
  autonomy.
\newblock \emph{International Research Journal of Engineering and Technology},
  3\penalty0 (9):\penalty0 369--372, 2016.

\bibitem[Jiang et~al.(2020)Jiang, Zhang, Zou, Ren, Lv, and
  Liu]{jiang2020learning}
Zhe Jiang, Yu~Zhang, Dongqing Zou, Jimmy Ren, Jiancheng Lv, and Yebin Liu.
\newblock Learning event-based motion deblurring.
\newblock In \emph{Proceedings of the IEEE/CVF Conference on Computer Vision
  and Pattern Recognition}, pp.\  3320--3329, 2020.

\bibitem[Jing et~al.(2021)Jing, Yang, Wang, Song, and Tao]{jing2021turning}
Yongcheng Jing, Yiding Yang, Xinchao Wang, Mingli Song, and Dacheng Tao.
\newblock Turning frequency to resolution: Video super-resolution via event
  cameras.
\newblock In \emph{Proceedings of the IEEE/CVF Conference on Computer Vision
  and Pattern Recognition}, pp.\  7772--7781, 2021.

\bibitem[Kim et~al.(2020)Kim, Song, Ye, and Baek]{kim2020pynet}
Byung-Hoon Kim, Joonyoung Song, Jong~Chul Ye, and JaeHyun Baek.
\newblock Pynet-ca: enhanced pynet with channel attention for end-to-end mobile
  image signal processing.
\newblock In \emph{European Conference on Computer Vision}, pp.\  202--212.
  Springer, 2020.

\bibitem[Kyung et~al.(2016)]{kyung2016theory}
Chong-Min Kyung et~al.
\newblock \emph{Theory and applications of smart cameras}.
\newblock Springer, 2016.

\bibitem[Lee \& Powers(2005)Lee and Powers]{lee2005comparison}
Yong-Keun Lee and John~M Powers.
\newblock Comparison of cie lab, ciede 2000, and din 99 color differences
  between various shades of resin composites.
\newblock \emph{International Journal of Prosthodontics}, 18\penalty0 (2),
  2005.

\bibitem[Li et~al.(2010)Li, Wei, and Zheng]{li2010design}
Zhaowen Li, Tingcun Wei, and Ran Zheng.
\newblock Design of black level calibration system for cmos image sensor.
\newblock In \emph{2010 International Conference on Computer Application and
  System Modeling (ICCASM 2010)}, volume~10, pp.\  V10--643. IEEE, 2010.

\bibitem[Liang et~al.(2024)Liang, Chen, Li, Lu, and Wang]{liang2024towards}
Guoqiang Liang, Kanghao Chen, Hangyu Li, Yunfan Lu, and Lin Wang.
\newblock Towards robust event-guided low-light image enhancement: A
  large-scale real-world event-image dataset and novel approach.
\newblock \emph{arXiv preprint arXiv:2404.00834}, 2024.

\bibitem[Liang et~al.(2023)Liang, Yang, Li, Duan, Xu, and
  Shi]{liang2023coherent}
Jinxiu Liang, Yixin Yang, Boyu Li, Peiqi Duan, Yong Xu, and Boxin Shi.
\newblock Coherent event guided low-light video enhancement.
\newblock In \emph{Proceedings of the IEEE/CVF International Conference on
  Computer Vision}, pp.\  10615--10625, 2023.

\bibitem[Liang et~al.(2021)Liang, Cai, Cao, and Zhang]{liang2021cameranet}
Zhetong Liang, Jianrui Cai, Zisheng Cao, and Lei Zhang.
\newblock Cameranet: A two-stage framework for effective camera isp learning.
\newblock \emph{IEEE Transactions on Image Processing}, 30:\penalty0
  2248--2262, 2021.

\bibitem[Liebe et~al.(1998)Liebe, Dennison, Hancock, Stirbl, and
  Pain]{liebe1998active}
Carl~C Liebe, Edwin~W Dennison, Bruce Hancock, Robert~C Stirbl, and Bedabrata
  Pain.
\newblock Active pixel sensor (aps) based star tracker.
\newblock In \emph{1998 IEEE Aerospace Conference Proceedings (Cat. No.
  98TH8339)}, volume~1, pp.\  119--127. IEEE, 1998.

\bibitem[Lin et~al.(2020)Lin, Zhang, Pan, Jiang, Zou, Wang, Chen, and
  Ren]{lin2020learning}
Songnan Lin, Jiawei Zhang, Jinshan Pan, Zhe Jiang, Dongqing Zou, Yongtian Wang,
  Jing Chen, and Jimmy Ren.
\newblock Learning event-driven video deblurring and interpolation.
\newblock In \emph{Computer Vision--ECCV 2020: 16th European Conference,
  Glasgow, UK, August 23--28, 2020, Proceedings, Part VIII 16}, pp.\  695--710.
  Springer, 2020.

\bibitem[Liu et~al.(2023)Liu, Wang, and Sun]{liu2023voxel}
Daikun Liu, Teng Wang, and Changyin Sun.
\newblock Voxel-based multi-scale transformer network for event stream
  processing.
\newblock \emph{IEEE Transactions on Circuits and Systems for Video
  Technology}, 2023.

\bibitem[Liu et~al.(2021)Liu, Lin, Cao, Hu, Wei, Zhang, Lin, and
  Guo]{liu2021swin}
Ze~Liu, Yutong Lin, Yue Cao, Han Hu, Yixuan Wei, Zheng Zhang, Stephen Lin, and
  Baining Guo.
\newblock Swin transformer: Hierarchical vision transformer using shifted
  windows.
\newblock In \emph{Proceedings of the IEEE/CVF international conference on
  computer vision}, pp.\  10012--10022, 2021.

\bibitem[Lu et~al.(2023{\natexlab{a}})Lu, Liang, and Wang]{lu2023self}
Yunfan Lu, Guoqiang Liang, and Lin Wang.
\newblock Self-supervised learning of event-guided video frame interpolation
  for rolling shutter frames.
\newblock \emph{arXiv preprint arXiv:2306.15507}, 2023{\natexlab{a}}.

\bibitem[Lu et~al.(2023{\natexlab{b}})Lu, Wang, Liu, Wang, and
  Wang]{lu2023learning}
Yunfan Lu, Zipeng Wang, Minjie Liu, Hongjian Wang, and Lin Wang.
\newblock Learning spatial-temporal implicit neural representations for
  event-guided video super-resolution.
\newblock In \emph{Proceedings of the IEEE/CVF Conference on Computer Vision
  and Pattern Recognition}, pp.\  1557--1567, 2023{\natexlab{b}}.

\bibitem[Lu et~al.(2024)Lu, Xu, Ma, Guo, and Xiong]{lu2024event}
Yunfan Lu, Yijie Xu, Wenzong Ma, Weiyu Guo, and Hui Xiong.
\newblock Event camera demosaicing via swin transformer and pixel-focus loss.
\newblock \emph{arXiv preprint arXiv:2404.02731}, 2024.

\bibitem[Luo et~al.(2001)Luo, Cui, and Rigg]{luo2001development}
M~Ronnier Luo, Guihua Cui, and Bryan Rigg.
\newblock The development of the cie 2000 colour-difference formula: Ciede2000.
\newblock \emph{Color Research \& Application: Endorsed by Inter-Society Color
  Council, The Colour Group (Great Britain), Canadian Society for Color, Color
  Science Association of Japan, Dutch Society for the Study of Color, The
  Swedish Colour Centre Foundation, Colour Society of Australia, Centre
  Fran{\c{c}}ais de la Couleur}, 26\penalty0 (5):\penalty0 340--350, 2001.

\bibitem[Mahy et~al.(1994)Mahy, Van~Eycken, and
  Oosterlinck]{mahy1994evaluation}
Marc Mahy, Luc Van~Eycken, and Andr{\'e} Oosterlinck.
\newblock Evaluation of uniform color spaces developed after the adoption of
  cielab and cieluv.
\newblock \emph{Color Research \& Application}, 19\penalty0 (2):\penalty0
  105--121, 1994.

\bibitem[McElvain \& Gish(2013)McElvain and Gish]{mcelvain2013camera}
Jon~S McElvain and Walter Gish.
\newblock Camera color correction using two-dimensional transforms.
\newblock In \emph{Color and Imaging Conference}, volume~21, pp.\  250--256.
  Society for Imaging Science and Technology, 2013.

\bibitem[Messikommer et~al.(2022)Messikommer, Georgoulis, Gehrig, Tulyakov,
  Erbach, Bochicchio, Li, and Scaramuzza]{messikommer2022multi}
Nico Messikommer, Stamatios Georgoulis, Daniel Gehrig, Stepan Tulyakov, Julius
  Erbach, Alfredo Bochicchio, Yuanyou Li, and Davide Scaramuzza.
\newblock Multi-bracket high dynamic range imaging with event cameras.
\newblock In \emph{Proceedings of the IEEE/CVF conference on computer vision
  and pattern recognition}, pp.\  547--557, 2022.

\bibitem[{MIPI Challenge 2024}(2024)]{mipi_2024}
{MIPI Challenge 2024}.
\newblock Mobile intelligent photography and imaging workshop 2024.
\newblock \url{https://mipi-challenge.org/MIPI2024/}, 2024.

\bibitem[Niklaus et~al.(2017)Niklaus, Mai, and Liu]{niklaus2017video}
Simon Niklaus, Long Mai, and Feng Liu.
\newblock Video frame interpolation via adaptive separable convolution.
\newblock In \emph{Proceedings of the IEEE international conference on computer
  vision}, pp.\  261--270, 2017.

\bibitem[Pan et~al.(2019)Pan, Scheerlinck, Yu, Hartley, Liu, and
  Dai]{pan2019bringing}
Liyuan Pan, Cedric Scheerlinck, Xin Yu, Richard Hartley, Miaomiao Liu, and
  Yuchao Dai.
\newblock Bringing a blurry frame alive at high frame-rate with an event
  camera.
\newblock In \emph{Proceedings of the IEEE/CVF Conference on Computer Vision
  and Pattern Recognition}, pp.\  6820--6829, 2019.

\bibitem[Paszke et~al.(2017)Paszke, Gross, Chintala, Chanan, Yang, DeVito, Lin,
  Desmaison, Antiga, and Lerer]{paszke2017automatic}
Adam Paszke, Sam Gross, Soumith Chintala, Gregory Chanan, Edward Yang, Zachary
  DeVito, Zeming Lin, Alban Desmaison, Luca Antiga, and Adam Lerer.
\newblock Automatic differentiation in pytorch.
\newblock 2017.

\bibitem[Poon \& Banerjee(2001)Poon and Banerjee]{poon2001contemporary}
T-C Poon and Partha~P Banerjee.
\newblock \emph{Contemporary optical image processing with MATLAB}.
\newblock Elsevier, 2001.

\bibitem[Qian et~al.(2017)Qian, Chen, Nikkanen, Kamarainen, and
  Matas]{Qian_2017_ICCV}
Yanlin Qian, Ke~Chen, Jarno Nikkanen, Joni-Kristian Kamarainen, and Jiri Matas.
\newblock Recurrent color constancy.
\newblock In \emph{Proceedings of the IEEE International Conference on Computer
  Vision (ICCV)}, Oct 2017.

\bibitem[Qian et~al.(2019)Qian, Kamarainen, Nikkanen, and
  Matas]{Qian_2019_CVPR}
Yanlin Qian, Joni-Kristian Kamarainen, Jarno Nikkanen, and Jiri Matas.
\newblock On finding gray pixels.
\newblock In \emph{Proceedings of the IEEE/CVF Conference on Computer Vision
  and Pattern Recognition (CVPR)}, June 2019.

\bibitem[Rainbow-Johnny-Johnny-Image-Processing-Lim(2022)]{quaddemosaic}
Rainbow-Johnny-Johnny-Image-Processing-Lim.
\newblock Quadbayer cfa modified gradient-based demosaicing, 2022.
\newblock URL
  \url{https://www.mathworks.com/matlabcentral/fileexchange/116085-quadbayer-cfa-modified-gradient-based-demosaicing}.
\newblock Accessed: 2024-06-01.

\bibitem[Ronneberger et~al.(2015)Ronneberger, Fischer, and
  Brox]{ronneberger2015u}
Olaf Ronneberger, Philipp Fischer, and Thomas Brox.
\newblock U-net: Convolutional networks for biomedical image segmentation.
\newblock In \emph{Medical image computing and computer-assisted
  intervention--MICCAI 2015: 18th international conference, Munich, Germany,
  October 5-9, 2015, proceedings, part III 18}, pp.\  234--241. Springer, 2015.

\bibitem[Russell et~al.(2008)Russell, Torralba, Murphy, and
  Freeman]{russell2008labelme}
Bryan~C Russell, Antonio Torralba, Kevin~P Murphy, and William~T Freeman.
\newblock Labelme: a database and web-based tool for image annotation.
\newblock \emph{International journal of computer vision}, 77:\penalty0
  157--173, 2008.

\bibitem[Scheerlinck et~al.(2019)Scheerlinck, Rebecq, Stoffregen, Barnes,
  Mahony, and Scaramuzza]{scheerlinck2019ced}
Cedric Scheerlinck, Henri Rebecq, Timo Stoffregen, Nick Barnes, Robert Mahony,
  and Davide Scaramuzza.
\newblock Ced: Color event camera dataset.
\newblock In \emph{Proceedings of the IEEE/CVF Conference on Computer Vision
  and Pattern Recognition Workshops}, pp.\  0--0, 2019.

\bibitem[Schwartz et~al.(2018)Schwartz, Giryes, and
  Bronstein]{schwartz2018deepisp}
Eli Schwartz, Raja Giryes, and Alex~M Bronstein.
\newblock Deepisp: Toward learning an end-to-end image processing pipeline.
\newblock \emph{IEEE Transactions on Image Processing}, 28\penalty0
  (2):\penalty0 912--923, 2018.

\bibitem[Shariff et~al.(2024)Shariff, Dilmaghani, Kielty, Moustafa, Lemley, and
  Corcoran]{shariff2024event}
Waseem Shariff, Mehdi~Sefidgar Dilmaghani, Paul Kielty, Mohamed Moustafa, Joe
  Lemley, and Peter Corcoran.
\newblock Event cameras in automotive sensing: A review.
\newblock \emph{IEEE Access}, 2024.

\bibitem[Shekhar~Tripathi et~al.(2022)Shekhar~Tripathi, Danelljan, Shukla,
  Timofte, and Van~Gool]{shekhar2022transform}
Ardhendu Shekhar~Tripathi, Martin Danelljan, Samarth Shukla, Radu Timofte, and
  Luc Van~Gool.
\newblock Transform your smartphone into a dslr camera: Learning the isp in the
  wild.
\newblock In \emph{European Conference on Computer Vision}, pp.\  625--641.
  Springer, 2022.

\bibitem[Song et~al.(2022)Song, Huang, and Bajaj]{song2022cir}
Chen Song, Qixing Huang, and Chandrajit Bajaj.
\newblock E-cir: Event-enhanced continuous intensity recovery.
\newblock In \emph{Proceedings of the IEEE/CVF Conference on Computer Vision
  and Pattern Recognition}, pp.\  7803--7812, 2022.

\bibitem[Tian et~al.(2002)Tian, Gledhill, Taylor, and Clarke]{tian2002colour}
Gui~Yun Tian, Duke Gledhill, David Taylor, and David Clarke.
\newblock Colour correction for panoramic imaging.
\newblock In \emph{Proceedings Sixth International Conference on Information
  Visualisation}, pp.\  483--488. IEEE, 2002.

\bibitem[Tulyakov et~al.(2021)Tulyakov, Gehrig, Georgoulis, Erbach, Gehrig, Li,
  and Scaramuzza]{tulyakov2021time}
Stepan Tulyakov, Daniel Gehrig, Stamatios Georgoulis, Julius Erbach, Mathias
  Gehrig, Yuanyou Li, and Davide Scaramuzza.
\newblock Time lens: Event-based video frame interpolation.
\newblock In \emph{Proceedings of the IEEE/CVF conference on computer vision
  and pattern recognition}, pp.\  16155--16164, 2021.

\bibitem[Tulyakov et~al.(2022)Tulyakov, Bochicchio, Gehrig, Georgoulis, Li, and
  Scaramuzza]{tulyakov2022time}
Stepan Tulyakov, Alfredo Bochicchio, Daniel Gehrig, Stamatios Georgoulis,
  Yuanyou Li, and Davide Scaramuzza.
\newblock Time lens++: Event-based frame interpolation with parametric
  non-linear flow and multi-scale fusion.
\newblock In \emph{Proceedings of the IEEE/CVF Conference on Computer Vision
  and Pattern Recognition}, pp.\  17755--17764, 2022.

\bibitem[Wang et~al.(2020{\natexlab{a}})Wang, He, Yu, Xia, and
  Yang]{wang2020event}
Bishan Wang, Jingwei He, Lei Yu, Gui-Song Xia, and Wen Yang.
\newblock Event enhanced high-quality image recovery.
\newblock In \emph{Computer Vision--ECCV 2020: 16th European Conference,
  Glasgow, UK, August 23--28, 2020, Proceedings, Part XIII 16}, pp.\  155--171.
  Springer, 2020{\natexlab{a}}.

\bibitem[Wang et~al.(2020{\natexlab{b}})Wang, Wu, Yuan, and
  Gao]{wang2020experiment}
Wencheng Wang, Xiaojin Wu, Xiaohui Yuan, and Zairui Gao.
\newblock An experiment-based review of low-light image enhancement methods.
\newblock \emph{Ieee Access}, 8:\penalty0 87884--87917, 2020{\natexlab{b}}.

\bibitem[Weng et~al.(2005)Weng, Chen, and Fuh]{weng2005novel}
Ching-Chih Weng, Homer Chen, and Chiou-Shann Fuh.
\newblock A novel automatic white balance method for digital still cameras.
\newblock In \emph{2005 IEEE International Symposium on Circuits and Systems
  (ISCAS)}, pp.\  3801--3804. IEEE, 2005.

\bibitem[Xiaopeng et~al.(2024)Xiaopeng, Zhaoyuan, Cien, Chen, Lei, and
  Lei]{xiaopeng2024hdr}
Li~Xiaopeng, Zeng Zhaoyuan, Fan Cien, Zhao Chen, Deng Lei, and Yu~Lei.
\newblock Hdr imaging for dynamic scenes with events.
\newblock \emph{arXiv preprint arXiv:2404.03210}, 2024.

\bibitem[Xing et~al.(2021)Xing, Qian, and Chen]{xing2021invertible}
Yazhou Xing, Zian Qian, and Qifeng Chen.
\newblock Invertible image signal processing.
\newblock In \emph{Proceedings of the IEEE/CVF Conference on Computer Vision
  and Pattern Recognition}, pp.\  6287--6296, 2021.

\bibitem[Xu et~al.(2021)Xu, Yu, Wang, Yang, Xia, Jia, Qiao, and
  Liu]{xu2021motion}
Fang Xu, Lei Yu, Bishan Wang, Wen Yang, Gui-Song Xia, Xu~Jia, Zhendong Qiao,
  and Jianzhuang Liu.
\newblock Motion deblurring with real events.
\newblock In \emph{Proceedings of the IEEE/CVF International Conference on
  Computer Vision}, pp.\  2583--2592, 2021.

\bibitem[Xu et~al.(2023)Xu, Hua, Zhang, Yu, and Qiao]{xu2023seeing}
Lexuan Xu, Guang Hua, Haijian Zhang, Lei Yu, and Ning Qiao.
\newblock " seeing" electric network frequency from events.
\newblock In \emph{Proceedings of the IEEE/CVF Conference on Computer Vision
  and Pattern Recognition}, pp.\  18022--18031, 2023.

\bibitem[Yang et~al.(2022)Yang, Yang, Jiang, Li, Feng, Zhou, Sun, Zhu, Loy, Gu,
  et~al.]{yang2022mipi}
Qingyu Yang, Guang Yang, Jun Jiang, Chongyi Li, Ruicheng Feng, Shangchen Zhou,
  Wenxiu Sun, Qingpeng Zhu, Chen~Change Loy, Jinwei Gu, et~al.
\newblock Mipi 2022 challenge on quad-bayer re-mosaic: Dataset and report.
\newblock In \emph{European Conference on Computer Vision}, pp.\  21--35.
  Springer, 2022.

\bibitem[Yaqi et~al.(2024)Yaqi, Zhihao, Xiaofeng, Jimmy~S., Xiaoming,
  Zongsheng, Chongyi, Shangcheng, Ruicheng, Yuekun, Peiqing, Chen~Change,
  et~al.]{hybridevs2024mipi3}
Wu~Yaqi, Fan Zhihao, Chu Xiaofeng, Ren Jimmy~S., Li~Xiaoming, Yue Zongsheng,
  Li~Chongyi, Zhou Shangcheng, Feng Ruicheng, Dai Yuekun, Yang Peiqing, Loy
  Chen~Change, et~al.
\newblock Mipi 2024 challenge on demosaic for hybridevs camera: Methods and
  results.
\newblock In \emph{Proceedings of the IEEE/CVF Conference on Computer Vision
  and Pattern Recognition}, 2024.

\bibitem[Yuan et~al.(2007)Yuan, Sun, Quan, and Shum]{yuan2007image}
Lu~Yuan, Jian Sun, Long Quan, and Heung-Yeung Shum.
\newblock Image deblurring with blurred/noisy image pairs.
\newblock In \emph{ACM SIGGRAPH 2007 papers}, pp.\  1--es. 2007.

\bibitem[Yunfan et~al.(2023)Yunfan, Liang, and Wang]{yunfan2023uniinr}
LU~Yunfan, Guoqiang Liang, and Lin Wang.
\newblock Uniinr: Unifying spatial-temporal inr for rs video correction,
  deblur, and interpolation with an event camera.
\newblock 2023.

\bibitem[Zhang et~al.(2022)Zhang, Ren, Luo, Lai, Stenger, Yang, and
  Li]{zhang2022deep}
Kaihao Zhang, Wenqi Ren, Wenhan Luo, Wei-Sheng Lai, Bj{\"o}rn Stenger,
  Ming-Hsuan Yang, and Hongdong Li.
\newblock Deep image deblurring: A survey.
\newblock \emph{International Journal of Computer Vision}, 130\penalty0
  (9):\penalty0 2103--2130, 2022.

\bibitem[Zhang \& Yu(2022)Zhang and Yu]{zhang2022unifying}
Xiang Zhang and Lei Yu.
\newblock Unifying motion deblurring and frame interpolation with events.
\newblock In \emph{Proceedings of the IEEE/CVF Conference on Computer Vision
  and Pattern Recognition}, pp.\  17765--17774, 2022.

\bibitem[Zhou et~al.(2022)Zhou, Duan, Ma, and Shi]{zhou2022evunroll}
Xinyu Zhou, Peiqi Duan, Yi~Ma, and Boxin Shi.
\newblock Evunroll: Neuromorphic events based rolling shutter image correction.
\newblock In \emph{Proceedings of the IEEE/CVF Conference on Computer Vision
  and Pattern Recognition}, pp.\  17775--17784, 2022.

\bibitem[Zhu et~al.(2018)Zhu, Wen, Bian, Ling, and Hu]{zhu2018vision}
Pengfei Zhu, Longyin Wen, Xiao Bian, Haibin Ling, and Qinghua Hu.
\newblock Vision meets drones: A challenge.
\newblock \emph{arXiv preprint arXiv:1804.07437}, 2018.

\end{thebibliography}
\bibliographystyle{iclr2025_conference}
}

\end{document}